\documentclass[sigconf]{acmart}

\AtBeginDocument{%
  \providecommand\BibTeX{{%
    \normalfont B\kern-0.5em{\scshape i\kern-0.25em b}\kern-0.8em\TeX}}}

\setcopyright{acmcopyright}
\copyrightyear{2023}
\acmYear{2023}
\acmDOI{XXXXXXX.XXXXXXX}

\acmConference[KDD DSHealth 2023]{Workshop on Applied Data Science for Healthcare: Applications and New Frontiers of Generative Models for Healthcare}{August 07,
  2023}{Long Beach, CA}
\begin{document}

\title[Model Performance Comparison for Predicting Drug-Review Satisfaction]{Bio+Clinical BERT, BERT Base, and CNN Performance Comparison for Predicting Drug-Review Satisfaction}

\author{Yue Ling}
\email{lingyue@berkeley.edu}
\orcid{0000-0002-8549-3996}
\affiliation{%
  \institution{University of California, Berkeley}
  \country{USA}
}

\renewcommand{\shortauthors}{Ling}

\begin{abstract}
  The objective of this study is to develop natural language processing (NLP) models that can analyze patients' drug reviews and accurately classify their satisfaction levels as positive, neutral, or negative. Such models would reduce the workload of healthcare professionals and provide greater insight into patients' quality of life, which is a critical indicator of treatment effectiveness.
  
  To achieve this, we implemented and evaluated several classification models, including a BERT base model, Bio+Clinical BERT, and a simpler CNN. Results indicate that the medical domain-specific Bio+Clinical BERT model significantly outperformed the general domain base BERT model, achieving macro f1 and recall score improvement of 11\%, as shown in Table \ref{table:model_perf}. Future research could explore how to capitalize on the specific strengths of each model. Bio+Clinical BERT excels in overall performance, particularly with medical jargon, while the simpler CNN demonstrates the ability to identify crucial words and accurately classify sentiment in texts with conflicting sentiments.
\end{abstract}



\received{10 June 2023}
\received[accepted]{7 July 2023}

\maketitle

\section{Introduction}

Healthcare clinics possess extensive records of patients' medical information, encompassing clinical notes, test results, survey responses, and drug reviews. Extracting valuable insights from these records regarding patients' responses to treatments often requires laborious manual reviews of textual data. However, automated inference techniques can offer valuable insights for guiding future treatment options.

Understanding patients' satisfaction with drugs is crucial as it can supplement established quality-of-life metrics and guide pharmaceutical research in identifying treatment-effective targets. Additionally, patients' drug reviews play a significant role in advocating for broader coverage of effective drugs by government and insurance companies.

Past work on health-related sentiment classification revolves around using a bag of words techniques like lexical matching or word frequencies, or shallow machine learning models with static or low-context embeddings \cite{Yadav:2018, Sokolova:2011, Pilan:2020, kotelnikova:2022}. While sentiment can be expressed in subtle and nuanced ways, more complex models have performed better in detecting sentiment in other domains \cite{Xiao:2018, tang:2015}. A more recent deep learning approach utilized BERT followed by a bidirectional LSTM architecture to achieve the highest drug review sentiment classification \cite{colon:2020}.

In recent years, transformers have significantly advanced NLP tasks by capturing contextual information from source text. Large pre-trained language models also facilitate handling complex text classification problems with limited datasets. This paper investigates the effectiveness of pre-trained language models, including BERT base and Bio+Clinical BERT, in classifying drug reviewers' treatment sentiments. We compare these models to a simpler CNN approach, aiming to gain insights into transformers and domain-specific pre-training for sentiment classification.

\section{Related Works}
Predicting drug satisfaction is challenging due to the diverse range of experiences and the complex nature of human sentiments. Negative ratings may stem from various factors, including persistent symptoms, severe side effects, and financial burdens, among others. Conversely, positive ratings can be influenced by factors such as effectiveness, absence of side effects, and affordability.

\citet{Sokolova:2011} employed basic machine learning models such as Naive Bayes, Decision Tree, K-nearest neighbor, and support vector machines with bag-of-word representations and human-annotation for drug review sentiment analysis. However, human annotation is an expensive process, whether it is used for standalone predictions or to prepare the data needed for training  \cite{Sokolova:2011}. 

Similarly, \citet{Yadav:2018} have also explored sentiments regarding the medical conditions of social media users based on users' self-narrated texts. In this work, \citet{Yadav:2018} used Google News pre-trained word embeddings with CNN to predict medical conditions (disease recovery/deterioration) and medication efficacy. Pre-trained embeddings fed into a downstream CNN architecture were able to highlight differentiating words that segment particular medical conditions. They were able to achieve f1-scores more than 20\% higher than their baselines' on predicting medical conditions and 7\% better on predicting medication efficacy. 

\citet{Safaya:2020} utilized a pre-trained BERT model combined with a CNN to identify offensive language on social media. In this case, the authors fed the last four hidden states of BERT into 160 convolutional filters of 5 different sizes to fine-tune their model. This BERT fine-tuned model outperformed a CNN alone by 5\% macro f1 score and the baseline model of an SVM applied to TF-IDF vectors by 9\%  macro f1 score. Their best model was able to leverage the BERT pre-trained context-rich embeddings with CNN filters to better discern offensive words. CNNs are often effective in sentiment analysis tasks because they focus on local contexts, such as phrase patterns that capture which adjectives modify which nouns, as well as negation.

\citet{Pilan:2020}  used hospital discharge summaries to find the presence of Syncope (fainting symptoms). The authors were able to get f1 scores 7\% higher than the baseline model with lexical matching to the "syncope" or "synkope" (Norwegian) term. Their best-performing model used word2vec embeddings trained on their own medical-domain data and outperformed the baseline by 12\% f1 scores. This suggests that fine-tuning domain-specific embeddings can work even better than general language embeddings for specific downstream tasks. 

In their study, \citet{Punith:2021} utilized the UCI ML Drug Review dataset and employed a different data-splitting approach. They conducted sentiment classification using various language models, including BERT, XLNET, ConvBERT, and Bio+Clinical BERT. ConvBERT outperformed the other models, achieving an accuracy 8.3\% higher than the ELMO baseline model. The authors also observed a substantial improvement when using the medical domain pre-trained model over the general pre-training tasks.

\section{Data} \label{dataset}
The UCI ML Drug Review dataset \cite{UCI:2018} includes 215,063 drug reviews from Drugs.com, each labeled with a 10-star rating. Our goal is to predict these ratings based on the review text, enabling inference of overall patient satisfaction from similar narratives like patient surveys. To simplify the analysis and highlight clearer differences between high, middle, and low scores, we binned the original labels into 3 categories. This helps address potential variability in how individuals assign similar scores. Binning the ratings into more consistent buckets allows us to focus on clearly misclassified instances that show substantial discrepancies, like those labeled as highly positive but predicted as negative, or vice versa. This approach avoids ambiguity in distinguishing between neutral compared to positive or negative sentiments, making our analysis more focused and comprehensive. The decision to bin the labels is driven by the practical application of the model, identifying the general positive or negative response to treatment options rather than fine-grained scores. This approach allows for determining whether the drug is effective for patients or if adjustments are needed. Binning the labels, as depicted in Table \ref{table:data}, results in more balanced data compared to the original 10 labels.

\begin{table}[h!]
\centering
\caption{Data Description}
\begin{tabular}{lccc} 
\toprule
Segment & Negative & Neutral & Positive \\
\midrule
Train & 40075 & 42702 & 78520 \\
Test & 13497 & 14076 & 26193 \\
\midrule
Binning & $\leq 4$ & $\geq 5$, $\leq 8$ & $\geq 9$ \\
\bottomrule
\end{tabular}
\label{table:data}
\end{table}

\section{Method and Experiments}
For this task, we compare several classification models using the base-cased general-purpose BERT model, Bio+Clinical BERT, a CNN applied to pre-trained word2vec embeddings. \cite{Alsentzer:2019, Devlin:2018, Mikolov:2013} We expect BERT to outperform CNN due to its ability to capture the longer-range context in drug reviews. BERT can incorporate the full narrative of reviewers discussing multiple experiences with a drug, whereas CNNs are limited to shorter-phrase patterns. Furthermore, Bio+Clinical BERT, specifically trained on medical text, is expected to outperform general domain BERT by recognizing domain-specific medical jargon. Our experiments compare the following models, each with a maximum token length of 128:

1. Baseline BERT base: pre-trained BERT embeddings, and pass the CLS token vector to a hidden dense layer of size 100, global max pooling, and a classification layer. We do not fine-tune.

2. CNN-Word2Vec: Pre-trained Word2Vec embeddings passed into a CNN with 50 or 100 filters of 1-5 tokens, followed by a hidden layer of size 100, global max pooling, and a classification layer. Converged at 18 epochs. 

3. BERT base: Pre-trained BERT model with fine-tuning on the last four layers. CLS token passed to a size 100 dense layer, global max pooling, and a classification layer. Converged at 8 epochs.

4. Bio+Clinical BERT: Pre-trained Bio+Clinical BERT model with fine-tuning on the last four layers. CLS token passed to a size 100 dense layer, global max pooling, and a classification layer. Converged at 11 epochs.

\section{Results and Discussions}

Consistent with similar research, we show macro precision, macro recall, and  macro f1 scores to evaluate model performance. The medical community commonly uses recall as the primary evaluation metric for predictions related to medical conditions and treatments, especially when diagnosing diseases due to the significant risk associated with missing cases requiring attention. In our case of detecting treatment dissatisfaction (the negative sentiment class), recall is also important since medical providers may want to further investigate and identify needed changes. 

The main results are shown in Table \ref{table:model_perf} and summarized below. As we expected, Bio+Clinical BERT classification model performs best, outperforming the more general domain BERT baseline model by 11\% in recall and f1 score. It also performs better than a CNN on its own by 4\% recall and 5\% f1 score. Compared to BERT base, Bio+Clinical BERT performs better by 1\% f1 score. Bio+Clinical BERT allows the model to leverage the domain-trained Bio+Clinical BERT embeddings but also tunes the [CLS] token to the specific task of predicting medicine review scores. 

\begin{table}[h!]
\centering
\caption{Model Performance on Test Data}
\begin{tabular}{lccc}
\toprule
Model & Precision* & Recall* & F1*\\
\midrule
Baseline BERT base & 0.72 & 0.69 & 0.70 \\
CNN-Word2Vec & 0.76 & 0.76 & 0.76 \\
BERT base & 0.80 & 0.80 & 0.80 \\
Bio+Clinical BERT & \textbf{0.81} & \textbf{0.80} & \textbf{0.81} \\
\bottomrule
\multicolumn{3}{l}{\footnotesize{* macro scores}}
\end{tabular}
\label{table:model_perf}
\end{table}

\subsection{Analysis of Misclassifications}
\label{sample}

To understand the strengths and performance differences among the models, we conducted manual reviews of misclassified examples. Specifically, we compared the misclassifications of the best-performing model (Bio+Clinical BERT) with other model options to identify patterns in correctly classifying different types of text. In the Appendix, we provide tables with example reviews that illustrate these patterns. We conducted a comparison of misclassified review scores between models, specifically focusing on cases where the actual score was 2 but predicted as 0, or vice versa. These instances of misclassification served as crucial indicators, highlighting the specific areas where one model demonstrated superior performance over the other. 

1. \textbf{Mislabeled examples}: Some reviews contradict the assigned numeric scores, indicating potential misinterpretation of the rating scale by reviewers. This common challenge in user-collected datasets from public platforms leads to misclassifications across all models. \emph{ Refer to \ref{table:Mislabeled example} for example reviews}.

2. \textbf{Contradictory language}: Reviews with seemingly contradictory sentiments about a drug are hard for all models to classify. From the sample of drug reviews that all of the models misclassified, another clearest trend was that they often contain both positive and negative sentiment comments in the same review. Reviewers sometimes talk about both the benefits and downsides of a drug and put a subtle emphasis on one or the other (e.g. by talking about the one that ultimately swayed their opinion last). \emph{Refer to \ref{table:Contradictory language} for example reviews}.

3. \textbf{Non-domain sentiment statements}: There were some examples that the simple CNN-only model actually did better on than Bio+Clinical BERT. Those were often examples in which the main sentiment statement didn’t have much if any medical jargon, but said something generic like “overall I’m happy with it” or “well worth the payoff”. These reviews might have also made comments about the trade-offs that went in the other direction of their main score, which the BERT models may have focused too much on. Bio+Clinical BERT appears to have more emphasis on comments involving medical terms, which aren’t always the main overall sentiment. \emph{Refer to \ref{table:Non-domain sentiment statements} for example reviews}.

4. \textbf{Medical-domain sentiment statements}: The use of medical slang, medicine names, and detailed disease symptoms in reviews is more accurately classified using Bio+Clinical BERT compared to base-cased BERT. This may be due to the domain-specific focus of the embeddings used in Bio+Clinical BERT, which allows for a better understanding of the underlying patient's treatment symptoms and associated sentiments. However, in some rare instances, base-cased BERT may predict better the overall structure of the review and pick up the overall sentiment despite the [UNK] tokens. It should be noted that Bio+Clinical BERT appears to place more emphasis on medical terms. \emph{Refer to \ref{table:Medical-domain sentiment statements1} and \ref{table:Medical-domain sentiment statements2} for example reviews}.

\section{Conclusion}

Correctly determining medicine satisfaction is a crucial task for devising effective treatment plans, and it is especially important to identify patients who are dissatisfied with their medication to prevent lasting side effects or impede full recovery. In this work, the best model, Bio+Clinical BERT, successfully addresses this task by accurately classifying patients’ drug review sentiment as positive, neutral, or negative, outperforming the baseline model by 11\% in f1 score.

While impressive, there is room for improvement in the model's performance. The CNN outperforms in cases where reviewers provide tangential information, capturing sentiment by filtering noise and identifying prominent phrases. Furthermore, CNN trains faster than BERT. Future work may involve combining these models to make multiple score predictions, enabling medical providers to identify patients who require further investigation.

\begin{acks}
I am grateful to Dr. Natalie Ahn for her outstanding mentorship, diligent review of my work, and valuable guidance throughout the project. Her constructive feedback and support have been invaluable.
\end{acks}

\bibliographystyle{ACM-Reference-Format}
\bibliography{references.bib}


\begin{thebibliography}{13}


\ifx \showCODEN    \undefined \def \showCODEN     #1{\unskip}     \fi
\ifx \showDOI      \undefined \def \showDOI       #1{#1}\fi
\ifx \showISBNx    \undefined \def \showISBNx     #1{\unskip}     \fi
\ifx \showISBNxiii \undefined \def \showISBNxiii  #1{\unskip}     \fi
\ifx \showISSN     \undefined \def \showISSN      #1{\unskip}     \fi
\ifx \showLCCN     \undefined \def \showLCCN      #1{\unskip}     \fi
\ifx \shownote     \undefined \def \shownote      #1{#1}          \fi
\ifx \showarticletitle \undefined \def \showarticletitle #1{#1}   \fi
\ifx \showURL      \undefined \def \showURL       {\relax}        \fi
\providecommand\bibfield[2]{#2}
\providecommand\bibinfo[2]{#2}
\providecommand\natexlab[1]{#1}
\providecommand\showeprint[2][]{arXiv:#2}

\bibitem[Alsentzer et~al\mbox{.}(2019)]%
        {Alsentzer:2019}
\bibfield{author}{\bibinfo{person}{Emily Alsentzer}, \bibinfo{person}{John~R
  Murphy}, \bibinfo{person}{Willie Boag}, \bibinfo{person}{Wei-Hung Weng},
  \bibinfo{person}{Di Jin}, \bibinfo{person}{Tristan Naumann}, {and}
  \bibinfo{person}{Matthew McDermott}.} \bibinfo{year}{2019}\natexlab{}.
\newblock \showarticletitle{Publicly available clinical BERT embeddings}.
\newblock \bibinfo{journal}{\emph{arXiv preprint arXiv:1904.03323}}
  (\bibinfo{year}{2019}).
\newblock


\bibitem[Col{\'o}n-Ruiz and Segura-Bedmar(2020)]%
        {colon:2020}
\bibfield{author}{\bibinfo{person}{Crist{\'o}bal Col{\'o}n-Ruiz} {and}
  \bibinfo{person}{Isabel Segura-Bedmar}.} \bibinfo{year}{2020}\natexlab{}.
\newblock \showarticletitle{Comparing deep learning architectures for sentiment
  analysis on drug reviews}.
\newblock \bibinfo{journal}{\emph{Journal of Biomedical Informatics}}
  \bibinfo{volume}{110} (\bibinfo{year}{2020}), \bibinfo{pages}{103539}.
\newblock


\bibitem[Devlin et~al\mbox{.}(2018)]%
        {Devlin:2018}
\bibfield{author}{\bibinfo{person}{Jacob Devlin}, \bibinfo{person}{Ming-Wei
  Chang}, \bibinfo{person}{Kenton Lee}, {and} \bibinfo{person}{Kristina
  Toutanova}.} \bibinfo{year}{2018}\natexlab{}.
\newblock \showarticletitle{Bert: Pre-training of deep bidirectional
  transformers for language understanding}.
\newblock \bibinfo{journal}{\emph{arXiv preprint arXiv:1810.04805}}
  (\bibinfo{year}{2018}).
\newblock


\bibitem[Gr\"{a}\ss{}er et~al\mbox{.}(2018)]%
        {UCI:2018}
\bibfield{author}{\bibinfo{person}{Felix Gr\"{a}\ss{}er},
  \bibinfo{person}{Surya Kallumadi}, \bibinfo{person}{Hagen Malberg}, {and}
  \bibinfo{person}{Sebastian Zaunseder}.} \bibinfo{year}{2018}\natexlab{}.
\newblock \showarticletitle{Aspect-Based Sentiment Analysis of Drug Reviews
  Applying Cross-Domain and Cross-Data Learning}. In
  \bibinfo{booktitle}{\emph{Proceedings of the 2018 International Conference on
  Digital Health}} (Lyon, France) \emph{(\bibinfo{series}{DH '18})}.
  \bibinfo{publisher}{Association for Computing Machinery},
  \bibinfo{address}{New York, NY, USA}, \bibinfo{pages}{121–125}.
\newblock
\showISBNx{9781450364935}
\urldef\tempurl%
\url{https://doi.org/10.1145/3194658.3194677}
\showDOI{\tempurl}


\bibitem[Kotelnikova et~al\mbox{.}(2022)]%
        {kotelnikova:2022}
\bibfield{author}{\bibinfo{person}{Anastasia Kotelnikova},
  \bibinfo{person}{Danil Paschenko}, \bibinfo{person}{Klavdiya Bochenina},
  {and} \bibinfo{person}{Evgeny Kotelnikov}.} \bibinfo{year}{2022}\natexlab{}.
\newblock \showarticletitle{Lexicon-based methods vs. BERT for text sentiment
  analysis}. In \bibinfo{booktitle}{\emph{Analysis of Images, Social Networks
  and Texts: 10th International Conference, AIST 2021, Tbilisi, Georgia,
  December 16--18, 2021, Revised Selected Papers}}. Springer,
  \bibinfo{pages}{71--83}.
\newblock


\bibitem[Mikolov et~al\mbox{.}(2013)]%
        {Mikolov:2013}
\bibfield{author}{\bibinfo{person}{Tomas Mikolov}, \bibinfo{person}{Kai Chen},
  \bibinfo{person}{Greg Corrado}, {and} \bibinfo{person}{Jeffrey Dean}.}
  \bibinfo{year}{2013}\natexlab{}.
\newblock \showarticletitle{Efficient estimation of word representations in
  vector space}.
\newblock \bibinfo{journal}{\emph{arXiv preprint arXiv:1301.3781}}
  (\bibinfo{year}{2013}).
\newblock


\bibitem[Pil{\'a}n et~al\mbox{.}(2020)]%
        {Pilan:2020}
\bibfield{author}{\bibinfo{person}{Ildik{\'o} Pil{\'a}n},
  \bibinfo{person}{P{\aa}l~H Brekke}, \bibinfo{person}{Fredrik~A Dahl},
  \bibinfo{person}{Tore Gundersen}, \bibinfo{person}{Haldor Husby},
  \bibinfo{person}{{\O}ystein Nytr{\o}}, {and} \bibinfo{person}{Lilja
  {\O}vrelid}.} \bibinfo{year}{2020}\natexlab{}.
\newblock \showarticletitle{Classification of Syncope Cases in Norwegian
  Medical Records}. In \bibinfo{booktitle}{\emph{Proceedings of the 3rd
  Clinical Natural Language Processing Workshop}}. \bibinfo{pages}{79--84}.
\newblock


\bibitem[Punith and Raketla(2021)]%
        {Punith:2021}
\bibfield{author}{\bibinfo{person}{NS Punith} {and} \bibinfo{person}{Krishna
  Raketla}.} \bibinfo{year}{2021}\natexlab{}.
\newblock \showarticletitle{Sentiment analysis of drug reviews using transfer
  learning}. In \bibinfo{booktitle}{\emph{2021 Third International Conference
  on Inventive Research in Computing Applications (ICIRCA)}}. IEEE,
  \bibinfo{pages}{1794--1799}.
\newblock


\bibitem[Safaya et~al\mbox{.}(2020)]%
        {Safaya:2020}
\bibfield{author}{\bibinfo{person}{Ali Safaya}, \bibinfo{person}{Moutasem
  Abdullatif}, {and} \bibinfo{person}{Deniz Yuret}.}
  \bibinfo{year}{2020}\natexlab{}.
\newblock \showarticletitle{Kuisail at semeval-2020 task 12: Bert-cnn for
  offensive speech identification in social media}. In
  \bibinfo{booktitle}{\emph{Proceedings of the Fourteenth Workshop on Semantic
  Evaluation}}. \bibinfo{pages}{2054--2059}.
\newblock


\bibitem[Sokolova and Bobicev(2011)]%
        {Sokolova:2011}
\bibfield{author}{\bibinfo{person}{Marina Sokolova} {and}
  \bibinfo{person}{Victoria Bobicev}.} \bibinfo{year}{2011}\natexlab{}.
\newblock \bibinfo{title}{Sentiments and Opinions in Health-related Web
  messages}.
\newblock , \bibinfo{numpages}{132--139}~pages.
\newblock


\bibitem[Tang et~al\mbox{.}(2015)]%
        {tang:2015}
\bibfield{author}{\bibinfo{person}{Duyu Tang}, \bibinfo{person}{Bing Qin},
  {and} \bibinfo{person}{Ting Liu}.} \bibinfo{year}{2015}\natexlab{}.
\newblock \showarticletitle{Document Modeling with Gated Recurrent Neural
  Network for Sentiment Classification}. In
  \bibinfo{booktitle}{\emph{Proceedings of the 2015 Conference on Empirical
  Methods in Natural Language Processing}}. \bibinfo{publisher}{Association for
  Computational Linguistics}, \bibinfo{address}{Lisbon, Portugal},
  \bibinfo{pages}{1422--1432}.
\newblock
\urldef\tempurl%
\url{https://doi.org/10.18653/v1/D15-1167}
\showDOI{\tempurl}


\bibitem[Xiao et~al\mbox{.}(2018)]%
        {Xiao:2018}
\bibfield{author}{\bibinfo{person}{Lizhong Xiao}, \bibinfo{person}{Guangzhong
  Wang}, {and} \bibinfo{person}{Yang Zuo}.} \bibinfo{year}{2018}\natexlab{}.
\newblock \showarticletitle{Research on Patent Text Classification Based on
  Word2Vec and LSTM}. In \bibinfo{booktitle}{\emph{2018 11th International
  Symposium on Computational Intelligence and Design (ISCID)}},
  Vol.~\bibinfo{volume}{01}. \bibinfo{pages}{71--74}.
\newblock
\urldef\tempurl%
\url{https://doi.org/10.1109/ISCID.2018.00023}
\showDOI{\tempurl}


\bibitem[Yadav et~al\mbox{.}(2018)]%
        {Yadav:2018}
\bibfield{author}{\bibinfo{person}{Shweta Yadav}, \bibinfo{person}{Asif Ekbal},
  \bibinfo{person}{Sriparna Saha}, {and} \bibinfo{person}{Pushpak
  Bhattacharyya}.} \bibinfo{year}{2018}\natexlab{}.
\newblock \showarticletitle{Medical sentiment analysis using social media:
  towards building a patient assisted system}. In
  \bibinfo{booktitle}{\emph{Proceedings of the Eleventh International
  Conference on Language Resources and Evaluation (LREC 2018)}}.
\newblock


\end{thebibliography}

\onecolumn

\appendix
\setcounter{table}{0} 
\renewcommand{\thetable}{A.\arabic{table}} 

\section*{Appendix}
\addcontentsline{toc}{section}{Appendix}

\begin{table}[htbp]
  \caption{Mislabeled examples: manually reviewed 37 examples to identify pattern}
  \centering
  \begin{tabular}{|p{9cm}|c|c|c|}
    \hline
    \textbf{Review Text} & \textbf{Label} & \textbf{Wrong} & \textbf{Correct}\\
    \hline
    \emph{Mislabeled example}: \newline
On my 4th shot. I have had a lot of muscle pain, especially in my left leg. Major spasms. Doctors think it is a slight bulging disk, but they have never seen this type pain with a bulging disk. Still not sure if it is associated with Repatha. 
 & 2 & All & None \\
    \hline
    \emph{Mislabeled example}: \newline
I started amitriptyline about 5 or 6 years ago. It was a miracle drug. But it has stopped working in the last 2 months. I039;ve been in the ER 5 times in 2 months…I am desperate for a new med before I loose my job!!!!
 & 2 & All & None \\
    \hline
    \emph{Mislabeled example}: \newline
Bottom line is the product DOES work and can heal your systems fully within 48hrs...only if you039;re up for the painful side effects. Honestly, if I had the option to use this again I probably would because it works CRAZY fast.
 & 0 & All & None \\
    \hline
    \emph{Mislabeled example}: \newline
3rd month, 3rd post I am happy to report that the getting 1-2 pimples every single day, HAS STOPPED! I am so happy…. Not to mention, I have always considered myself very emotionally aware/stable, but this pill has mellowed me out to a whole new level. Love it!
 & 0 & All & None \\
    \hline
  \end{tabular}
\label{table:Mislabeled example}
\end{table}

\begin{table}[htbp]
\caption{Contradictory language: manually reviewed 37 examples to identify pattern}
  \centering
  \begin{tabular}{|p{9cm}|c|c|c|}
    \hline
    \textbf{Review Text} & \textbf{Label} & \textbf{Wrong} & \textbf{Correct}\\
    \hline
    \emph{Contradictory language}: \newline
A few friends reccomended this brand and said it did not give them any side effects. I was sold on that alone. A couple of weeks in, I started to get horrible cystic acne on my chin…I almost wanted to get off the pill because of it. But I loved that I did not have any other side effects. Finally, half way through my 3rd pack the cysts were gone. All in all, it's a great pill besides the cysts I got.
 & 2 & All & None \\
    \hline
    \emph{Contradictory language}: \newline
I read the info sheet that was given to me when I got Drysol….The first night I used it it burned to the point where after an hour I had to wash it off just so I could get to sleep. Even then it burned but it wasn't unbearable. The next night same thing. I'm liking the product as it helps but it isn't worth the pain.
 & 2 & All & None \\
    \hline
    \emph{Contradictory language}: \newline
I used the Obagi Nu Derm System for sun spots in 2007 and had amazing results… however because my face had a burning sensation while I was on my computer. Today 8 years later I still get the burning face sensation when I'm on the computer or driving my car during the day even with sunscreen on.
 & 0 & All & None \\
    \hline
    \emph{Contradictory language}: \newline
Years ago I was on the older formula of Fentanyl and it worked wonders with minimal dosage \& breakthrough meds. After the companies reformulated the drug it isn't nearly as good as it used to be \& has many more side effects, brand depending. Mylan works best for me as far as fewer side effects but none work for the 72 hours they are supposed to. 
 & 0 & All & None \\
    \hline
    
  \end{tabular}
\label{table:Contradictory language}
\end{table}

\begin{table}[htbp]
\caption{Non-domain sentiment statements: manually reviewed 13 examples to identify pattern}
  \centering
  \begin{tabular}{|p{9cm}|c|c|c|}
    \hline
    \textbf{Review Text} & \textbf{Label} & \textbf{Wrong} & \textbf{Correct}\\
    \hline
    \emph{Non-domain sentiment statements}: \newline 
I am a 67 year old female who was diagnosed with A-Fib and put on Rythmol in 1996…. The doctor was about to do an ablation but decided to try Rythmol first.  It put me back in rhythm and I did not have to have the ablation.  I take 225 mg of propafenone every 8 hours (6, 2, and 10).  It is a nuisance sometimes, but well worth the payoff. I sometimes even forget that I am a "heart patient".
 & 2 & Bio+Clinical BERT & CNN \\
    \hline
    \emph{Non-domain sentiment statements}: \newline 
I had no discharge, just itching inside and out the vejay. I went…and got the 1 day Monistat Ointment. Immediately I felt the ointment was working, because I felt the itching had stopped and I felt a mild burning sensation. I felt a little burning sensation on the outside of my vejay, very mild sensation. I wiped myself with tissue in cold water and went back to sleep. The next morning I had no symptoms. I would recommend Monistat 1 ointment to cure yeast infections fast.
 & 2 & Bio+Clinical BERT & CNN \\
    \hline
    \emph{Non-domain sentiment statements}: \newline 
I suffer from anxiety and depression.  Doc suggested this drug, crying daily, filled with terror. The only thing it did was stop the daily crying and provide an odd clarity. 
So many side effects, nausea, headache and terrible depression, anxiety and suicidal thoughts, diarrhea, constipation, ear ringing, withdrawn.
 & 0 & Bio+Clinical BERT & CNN \\
    \hline
    \emph{Non-domain sentiment statements}: \newline 
First off this medicine is waaaaay over priced. My doctor didn't have any samples but I went online and they offer a free ten day trial. I highly suggest you try it before you commit.  It did not make me fall asleep or remain asleep. I would place this right up there with a heavy dose of Tylenol pm's.
 & 0 & Bio+Clinical BERT & CNN \\
    \hline
    \emph{Non-domain sentiment statements}: \newline 
I am 56 year old female with Severe osteoporosis and osteoarthritus. My Ortho and Gyno Dr.said My bones were so frail I had no choice but take forteo.I started in 2015 and after a few months of just over all weakness and being sick all the time low BP, I went to my PC Dr. After blood work he said your immune System has crashed get off the forteo! I was home bound had pneumonia etc…. Over all not well feeling . I'm happy for the ones who can take it with no side effects. !
 & 0 & Bio+Clinical BERT & CNN \\
    \hline
    
  \end{tabular}
\label{table:Non-domain sentiment statements}
\end{table}

\begin{table}[htbp]
\caption{Medical-domain sentiment statements 1 of 2: manually reviewed 13 examples to identify pattern}
  \centering
  \begin{tabular}{|p{7cm}|c|c|c|}
    \hline
    \textbf{Review Text} & \textbf{Label} & \textbf{Wrong} & \textbf{Correct}\\
    \hline
    \emph{Medical-domain sentiment statements}: \newline
Many people said taking Valacyclovir or Valtrex only works if taken at first signs (the dreaded tingles), well obviously it was impossible to take it at the first sign for me since I had to call for an appt with the doc, go to the doc, get a prescription, go to the pharmacy \&amp; wait for it to be filled. By the time I had the meds on hand, it was day 2 of a horrible outbreak (7 blisters all over my mouth, filled with fluids and swollen lips) and even though the blisters had formed - I took my doses, \&amp; by the next day ALL the swelling was gone, most of them were barely noticeable, and the big ones were already stabbing over and less noticeable. Usually it takes 6 days to scab over \&amp; heal \&amp; I'm only on day 3.
 & 2 & Base-cased BERT & Bio+Clinical BERT \\
    \hline
    \emph{Medical-domain sentiment statements}: \newline
Taking this medication after 3 day I experienced watery bowel motion.
 & 0 & Base-cased BERT & Bio+Clinical BERT \\
    \hline
    \emph{Medical-domain sentiment statements}: \newline
I have been a nurse since 1984 and a paramedic since 1990. In New Orleans,in 1993, while working on an ambulance, I became nauseated and began vomiting. Went home,changed ,had someone drive me 2 ER @ a Hospital in Harahan(nice subdivision.) I expected 2 get the usual treatment of IV Saline and the SAFE anti-emetic Phenergan. I told the ER DOC I was VERY ALLERGIC 2 REGLAN.  He injects me with Inapsine(Droperidol) because he wants  to see  how it worked - even though people with allergy to Reglan should NEVER be given Inapsine. I had already been declared clinically dead 2 years prior in a car accident, But was revived. This So-called drug causes Irregular heartbeats. I thought I was going 2 die. It was much worse than actually dying.
 & 0 & Base-cased BERT & Bio+Clinical BERT \\
    \hline  
  \end{tabular}
\label{table:Medical-domain sentiment statements1}
\end{table}

\begin{table}[htbp]
\caption{Medical-domain sentiment statements 2 of 2: manually reviewed 13 examples to identify pattern}
  \centering
  \begin{tabular}{|p{7cm}|c|c|c|}
    \hline
    \textbf{Review Text} & \textbf{Label} & \textbf{Wrong} & \textbf{Correct}\\
    \hline

    \emph{Medical-domain sentiment statements}: \newline
Read all prescriptions profiles provided by pharmacy!  Have taken drug for 3+ years and by accident learned I hadve side effects to it.  I have experienced 80\% of all adverse reactions listed; the worse being constant copious amounts of choking mucus/phlegm from throat/sinus, asthmatic bronchitis, severe fatigue, etc.  Never in a million years would I have guessed it was Lisinopril, but just happened to be out of the medication for 3 days and woke up on the 3rd day without any of the choking phlegm in my throat \&amp; bronchial tract every morning for the past 3+ years.  Current Doc (who didn't initially prescribe drug) is thrilled to discover the culprit he's been searching for  2+ years!  Good-bye erroneous early-stage COPD diagnoses, I'm cured!
 & 0 & Base-cased BERT & Bio+Clinical BERT \\
    \hline
    \emph{Medical-domain sentiment statements}: \newline
I used to experience a lot of gas due to GERD but not since I've been taking Dexilant.
 & 2 & Bio+Clinical BERT & Base-cased BERT \\
    \hline  
  \end{tabular}
\label{table:Medical-domain sentiment statements2}
\end{table}

\end{document}